\documentclass{jfsma}
 
\usepackage[latin1]{inputenc}
\usepackage{stmaryrd}
\usepackage{amssymb}
\usepackage{latexsym,amsmath,amssymb,textcomp}
\usepackage{graphicx}
\usepackage{subfigure}
\usepackage{url}
\usepackage{lscape}
\usepackage{soul}
\usepackage{color}
\usepackage{tabularx}
\usepackage{todonotes}
\titre{Multi-agent model for risk prediction in surgery}

\auteur{Bruno Perez\up{a}}{bruno.perez@univ-fcomte.fr}
\auteur{Christophe Lang\up{a}}{christophe.lang@univ-fcomte.fr}
\auteur{Julien Henriet\up{a}}{julien.henriet@univ-fcomte.fr}
\auteur{Laurent Philippe\up{a}}{laurent.philippe@univ-fcomte.fr}

\institution{\up{a}%
  Institut FEMTO-ST/CNRS\\
  University of Bourgogne-Franche-Comt{\'e}, France}

\date{\small \today}

\begin{document}
	
\maketitle
\pagestyle{empty}
\thispagestyle{empty}

%

\motscles{multi-agent system, case-based reasoning, prediction,
  modeling, surgery}

\bigskip

\begin{abstract}
 Risk management resulting from the actions and
states of the different elements making up a
operating room is a major concern 
 during a surgical procedure.
Agent-based simulation shows an interest through its interaction concepts, interactivity and autonomy of different
simulator entities. We want in our study to implement a generator
of alerts to listen the evolution of different
settings applied to the  simulator of agents
(human fatigue, material efficiency,
infection rate ...). This article presents our model, its implementation and the first results obtained. It should be noted that this study also made it possible to identify several scientific obstacles, such as the integration of different levels of abstraction, the coupling of species, the coexistence of several scales in the same environment and the deduction of unpredictable alerts. Case-based reasoning (CBR) is a beginning of response relative to the last lock mentioned and will be discussed in this paper.
\end{abstract}

\keywords{multi-agent system, artificial intelligence, predictive,
   modeling, multiscale, surgery}

\section{Introduction}
In hospitals, surgery is practiced in increasingly innovative and efficient contexts. Instruments are connected and maintained with all the rigor imposed by hygiene and safety standards. All parameters are checked before and after the surgery  (identity and condition of the patient, surgical materials, surgical instruments, etc.) in such a way as to leave as little room as possible for unforeseen events. But even in this safe context, minimizing the risk of a serious adverse event remains one of the main concerns of practitioners.  Indeed, despite all the protocols implemented, the infectious risk for the bloc, even in the case of clean surgery, is 1\% according to the College of University of Infectious and  Tropical Diseases Academics (CUITD). Among  other failures include dosage errors, clumsiness of movement and human fatigue.


In this context, we aim to model the operating room environment and resources in order to predict and quantify patient exposure to risk. Our ambition is to create a simulator able to generate a large number of varied situations in order to interpolate and quantify the possible evolutions of a given situation.

Our study, which is oriented towards predictive models composed of non-deterministic entities, raises, among other things, the problem of determining alert thresholds in close connection with the acquisition of knowledge. Indeed, is it possible to optimize the definition of thresholds and at the same time to aggregate a dynamic knowledge base to our system?  
   A possible answer is the agentification of thresholds, enriched by the coupling between MAS and CBR (case based reasoning) which solves a target problem by analogy with previous knowledge (case base) that evolves. 
In this paper, we propose to explore the conceptual approaches related to our positioning. The literature specific to the MAS/CBR coupling and related to the predictive domain will be studied in the section \ref{sec:eda} devoted to the state of the art. Some definitions and reminders will be given in preamble in this section.  In the section \ref{sec:expe}, we will focus on the characterization of criticality, then on the interest of the MAS/CBR coupling in response to our problem: "risk predictability in a non-deterministic context". We will specify our position and our contributions applied to these problems in a theoretical context. We will present our first results in the section \ref{sec:result} and we will comment  them in the following section.
The conclusion will characterize the relevance of our choices and the perspectives to be considered.

\section{Definitions and Related Work}
\label{sec:eda}
Predictive systems aim to deliver new information extracted from simulations or past learning. They depend mainly on available data and on the environment. The era of big data is a major current lead in the production of the latter, which we want to be both exhaustive and reliable. The question arises when access to such big data is restricted. Our problem which enters in this case obliges us to consider other systems such as MAS coupled to CBR able to predict with little initial knowledge. In the rest of this section, we will recall the principles and definitions of these two paradigms.

Definitions from the literature characterize MAS as systems composed of autonomous entities that interact with each other according to certain rules in a specific environment. In ~\cite{ferber1998multi}, Ferber defines agents as autonomous software entities able to perceive  (by messages, data capture ...) in an environment closed to the outside world.  In this context, cognitive agents act with reflection and awareness of their environment composed of other entities, unlike reactive agents that react only to stimuli. Thus, in the case of complex problems such as the epidemiological and ecological analysis of infectious diseases, standard models based on differential equations very quickly become unsuitable due to too many parameters and are supplanted by the MAS \cite{roche2008multi}. We also observe the increasingly frequent integration of the agent paradigm in the reinforcement of machine learning \cite{kok2006collaborative} which cannot be dissociated from knowledge acquisition. It characterizes the autonomy of agents and can be implemented beforehand (innate knowledge) or acquired through experience. This second case relates to the capacity of the entity to explore a knowledge base \cite{shen2015elaboration}. Its performance depends mainly on the collection of information that enriches its learning. The CBR architecture, designed according to this principle, can be a solution to this requirement.

The notion of knowledge acquisition is at the heart of dynamic models enriched by experience. Our modeling of non-deterministic entities falls within this framework and the values of the solutions
are not necessarily known. It is therefore essential to have
a memory of previous cases and an ontology to structure them.
Case-Based Reasoning (CBR) is a paradigm of artificial intelligence based on the elaboration of solutions by analogy with past experiences and general knowledge in the field of application. It is widely used in 
medicine~\cite{choudhury2016survey}, in the industrial maintenance 
~\cite{camarillo2017cbr} or in the stock market analysis ~\cite{musto2015personalized}. To find a solution, the CBR needs a base of solved cases, the characterization of similarities and finally the knowledge of adaptation processes. Several books deal with knowledge acquisition, similarity or decision making with MAS~\cite{giannakis2016multi}. The different predictive approaches structure their modeling based on the analysis of past experiences. The nature of the data and the analysis tools distinguish these predictive concepts. Two main approaches are considered in the literature for the use of CBR in an MAS. In the first approach, the linkage between CBR and MAS integrates into the agent the ability to solve problems based on experiences extracted from a case base. In ~\cite{jubair2018survey} problems are solved locally with sometimes a collaborative approach \cite{ontanon2007learning}. In ~\cite{mahmoud2015context}, case based reasoning gives the multi-agent system the ability to access a structured database that speeds up data mining. However, this case-based search mode built for a specific problem shows its limitations on non-similar cases. Moreover, the CBR/MAS coupling brings a dynamic character to the CBR (static) whereas we would like a dynamic modeling enriched by experiments. Indeed, the multi-agent simulation remains at the heart of our work whose stake consists in "predicting in a non-deterministic context". CBR is therefore integrated as a source of active knowledge acquisition.   
The second approach proposes rather a decision support system in a multi-agent environment. One of the main objectives is to suggest possible responses to different contexts constrained by numerous
parameters. Once again, the interest of a case-by-case reasoning
lies in its ability to find solutions by analogy. These
can be ordered in a hierarchy (bargaining agent (BA),
expert agent (EA))~\cite{tolchinsky2006cbr} or organized in collaborative committees. The
system can explore its own case database, several merged databases, or independent databases. The search for similar cases integrates different principles of initiation and learning such as artificial neural networks~\cite{pinzon2011s}. These decision support tools that integrate greater adaptability in the acquisition of knowledge are one of the possibilities we have explored. This type of coupling
is, however, limited to case bases, while other types of
data such as traces are interesting sources of information. 
The different MAS coupled with case-based reasoning do not provide a complete answer to our problem of risk prediction in a non-deterministic context.  Our model must not only integrate the dynamic interaction with all the entities of its environment (human and material) but must also integrate the anticipation of non-deterministic situations through a multidimensional approach of the phenomena. 
Independently of the dynamic interaction with all the entities of its environment (human and material), our model must also integrate the anticipation of non-deterministic situations through a multidimensional approach of the phenomena.
We therefore propose a new architecture that addresses this issue in the following section. 
\vspace{-1ex}

\section{An MAS model for the operating room}
\label{sec:expe}
In this non-deterministic context, and in order to meet our objective (to predict human and material failures), we decided to model the operating room as a multi-agent system. This type of simulation opens up a large number of perspectives and makes it possible to highlight risks that have been minimized or ignored until now, and finally to make a contribution in terms of safety in the operating room.
UML (Unified Modeling Language) \cite{specificationuml} and AML (Agent Modeling Language) have been chosen as the modeling language to develop the architecture of our MAS. 

%
 
 The architecture of our model, which integrates the BDI  (Belief, Desire , Intention) paradigm, has five species of entities: Personal, Material, Infection, Patient (singleton), Alert (singleton). Figure~\ref{tab:states} gives a description of the state variables of these agents :
\vspace{-1ex}
  \begin{figure}
  \begin{tabular}{|l|l|l|}
    \hline
    {\small Agent} &{\small  Variables} & {\small Comment} \\
    \hline
    {\small Personal} & {\small \textit{intention}} & \begin{small}
    operate a patient in
    \end{small} \\
     &  & \begin{small}optimal conditions
     \end{small} \\
	 &  &  \begin{small} fail-safe   \end{small} \\ 
          & {\small \textit{desire}} &  \begin{small}   use resources \end{small} \\   
     &  & \begin{small}  human and material   \end{small} \\  
	 &  & \begin{small}(personal, material)    \end{small} \\  
          & {\small \textit{belief}} & \begin{small}   appropriate measures to take \end{small} \\   
	 &  & \begin{small} decision (monitoring,   \end{small} \\   
	 &  & \begin{small} alert)   \end{small} \\  
          &{\small \textit{tiredness}} & {\small fatigue rate (scale}  \\
	 &  & \begin{small}of 1 slightly tired     \end{small} \\    
	 &  & \begin{small} up to 5 sold out)   \end{small} \\    
          & {\small \textit{experience}} & \begin{small}   junior, senior \end{small} \\   
    \hline
   {\small  Material} & {\small \textit{function}} & \begin{small}    hardware functionality \end{small} \\ 
          & {\small \textit{mat\_tiredness}} & \begin{small}   material efficiency  \end{small} \\   
	 &  & \begin{small} (scale of 1 effective to    \end{small} \\   
	 &  & \begin{small}   3 Ineffective)  \end{small} \\  
          & {\small\textit{infected}}& \begin{small}   boolean \end{small} \\ 
    \hline
    {\small Infection} & {\small \textit{type}} & \begin{small}    type of infectious agent \end{small} \\  
 	 &  & \begin{small} (contaminant, resistant)   \end{small} \\     
          & {\small \textit{local}} & \begin{small}   has an impact on an area,  \end{small} \\  
 	 &  & \begin{small} in the operating room   \end{small} \\    
 	 &  &  \begin{small} or both    \end{small} \\ 	                         
          & {\small \textit{desire}} & \begin{small}   set according to \end{small} \\    
 	 &  & \begin{small}  of its kind (contaminant,    \end{small} \\        
 	 &  & \begin{small} resistant)  \end{small} \\  
          & {\small \textit{belief}}& \begin{small}    assess recovery \end{small} \\  
 	 &  & \begin{small} with the future host   \end{small} \\  

    \hline
   {\small  Patient} & {\small \textit{state}} &  \begin{small}   state of health \end{small} \\ 
          & {\small \textit{surgery\_type}} & \begin{small}    urgent, non-urgent, \end{small} \\   
 	 &  & \begin{small}  complexe, non complexe  \end{small} \\  
    \hline
    {\small Alert} & {\small \textit{intention}} & \begin{small}    prevent failure \end{small} \\ 
          & {\small \textit{desire}} & \begin{small}   early warning (before  \end{small} \\  
 	 &  & \begin{small}that the failure does not     \end{small} \\          
 	 &  & \begin{small}   produces) \end{small} \\   
          & {\small \textit{belief}} & \begin{small}   listen and follow \end{small} \\  
  	 &  & \begin{small} the evolution of the data   \end{small} \\            
  	 &  & \begin{small} influencing surgical   \end{small} \\  
  	 &  & {\small intervention}\\
          & {\small \textit{level}} & {\small warning signs}\\
    \hline
  \end{tabular}
  \caption{Simulated agent state variabless}
  \label{tab:states}
\end{figure}

On an experimental basis, we simulated the evolution of an
infectious site in parallel with human fatigue by calibrating simulation cycles to 30 seconds.

\paragraph{Fatigue modeling.}
Fatigue can be modeled using several data acquisition methods. In unconnected mode, data are extracted from statistical files or from a function. If it is a dynamic capture, connected sensors (bracelet for fatigue, electrochemical sensors for infection rate) as well as monitoring will provide the information. In our case, we have chosen to define the growth of fatigue from an exponential function (parameterizable by the user) because it applies to continuous phenomena and highlights the non-linear nature of fatigue. It is given by the relation : \\
\begin{Large}$f(t) = ae^{(k\times t)}$ \end{Large} ou \begin{Large} $a$\end{Large} is the initial value, \begin{Large} $k$ \end{Large} constant growth, and \begin{Large} $t$ \end{Large} the time variable. \ 
   In the second phase of the project, dynamic data capture will be preferred. Note that in our case we have treated the level of fatigue of a single agent knowing that each one  may have his own risk threshold.. \\

 \paragraph{Infection modeling.}


The two main types of major contamination that can be observed are, on the one hand, exogenous agents from the operating room and, on the other hand, endogenous agents.
Our system is able to simulate the evolution of the infection rate of exogenous hosts as a function of decontaminating agents and time. The model is initialized according to the following parameterizable values :

\begin{itemize}
	\item Number of susceptible host (healthy particle count): 495 ;
	\item Number of infected host (number of infected particles): 5 cfu (colony forming unit) / m $^{3}$;
	\item Number of resistant host (number of resistant particles).
\end{itemize}
The ratio of infected particles to healthy particles ($\dfrac{5}{495}$) indicates an operating site infection (OSI) rate of approximately 1 \%. This percentage, proposed by the medical team, appears in several studies such as \cite{alfonso2017analyzing}\cite{florentin2011construction}.     
The progression or regression of infection is related to the overlap between healthy and contaminated particles over time.

\paragraph{Simulation of fatigue coupled with infection.}
The coupling of the two failures provides a global vision of the risks in the operating room. It is then possible to define an alert level correlated to several risks. 
The result of the interaction between the entities (infection, fatigue) is presented in Part \ref{sec:result}.  \\

\paragraph{Agent Alert.} 
The cognitive agent Alert is central in our architecture and plays the role of a cognitive agent. :

\begin{itemize}
	\item centralizing (centralizing the different alert thresholds, etc.));
	\item adapter (manages a collective alert level);
	\item regulator (proposes possible solutions).
\end{itemize}

Our originality with respect to the approaches described in this section lies first of all in the fact that we have designed a multi-scale system of agents in which human (fatigue) and material (efficiency) failures as well as nosocomial diseases are modelled. Thus we have converged (centralized) types of data with different scales and metrics. On the other hand,
one of the keystones of our predictive modeling is formalized by the aggregation of the different levels specific to each indicator (state variable) listened to by the system. One of the resulting problems concerns : 
\begin{itemize}
	\item the definition;
	\item representation;
	\item measurement;
\end{itemize} 
of these collective thresholds. 

\begin{figure}[h!]
	\centering
	\includegraphics[scale=0.3]{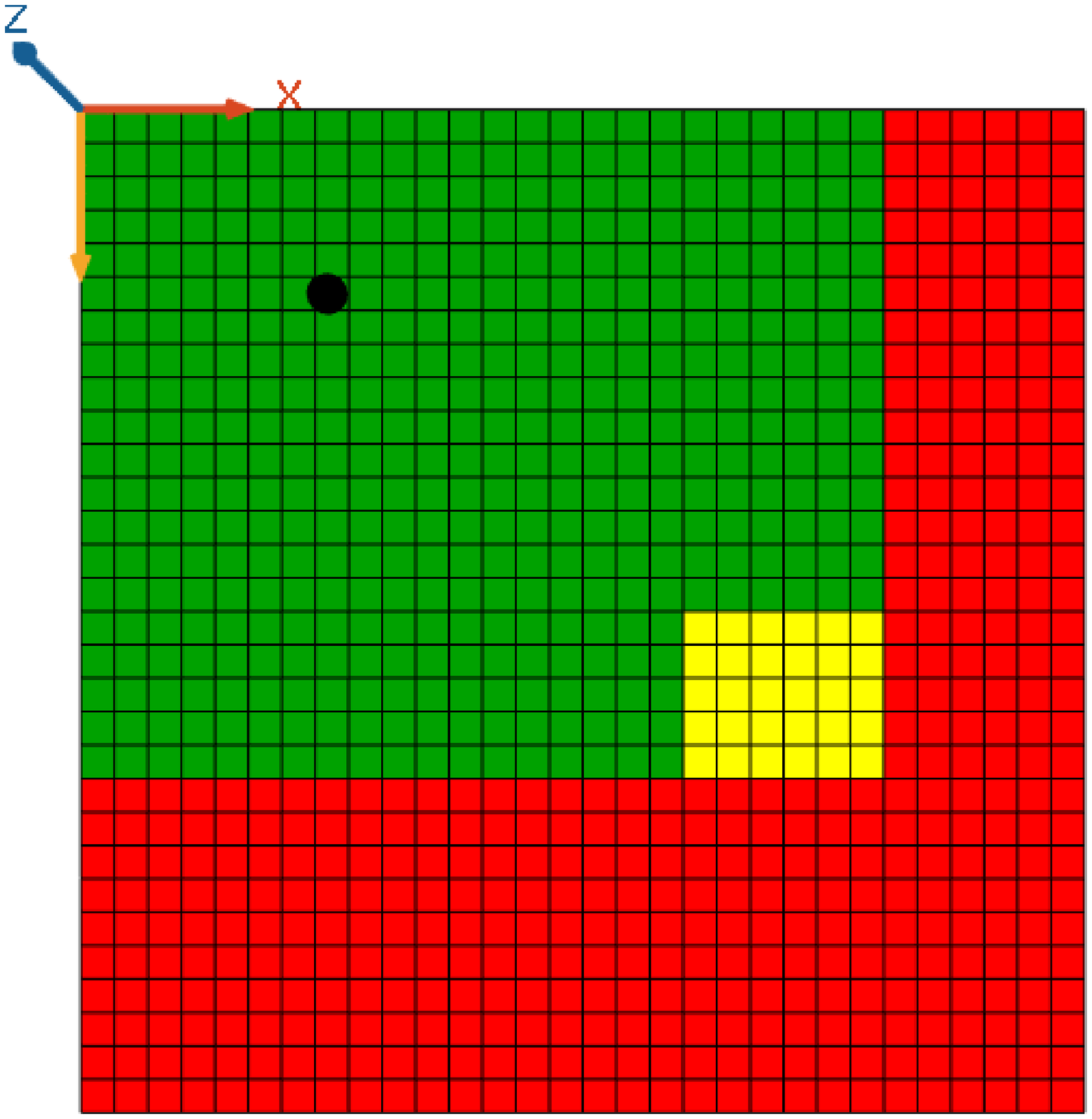}
	\caption{Identification of the level of risk between two variables}
	\label{position_var_2d}
\end{figure}

In response to this questioning, we propose a multidimensional approach. It is not a question of comparing trend curves or analysing dispersions, but rather of identifying a point in the affine plane (2D) or affine space (3D).
In the first case, 2 trends evolve in an affine space such that to each coordinate corresponds a cell (agent) of a grid able to interact with its environment (overlap recognition). The example of Figure \ref{position_var_2d} is a graphical representation of the criticality between two variables (infectious\_rate, human\_tiredness) belonging respectively to the species \textit{personal} and \textit{infection}. The evolution of these variables depends on the behaviors and capacities of the agents (BDI) but also on the interaction between these two species. Each point of coordinates (infectious\_rate, human\_tiredness) overlaid on a grid cell typed according to a criticality level. In our example, the position of the coordinate point (human infectivity, infectious rate) indicates an acceptable level of criticality (green). 

\begin{figure}[h!]
	\centering
	\includegraphics[scale=0.2]{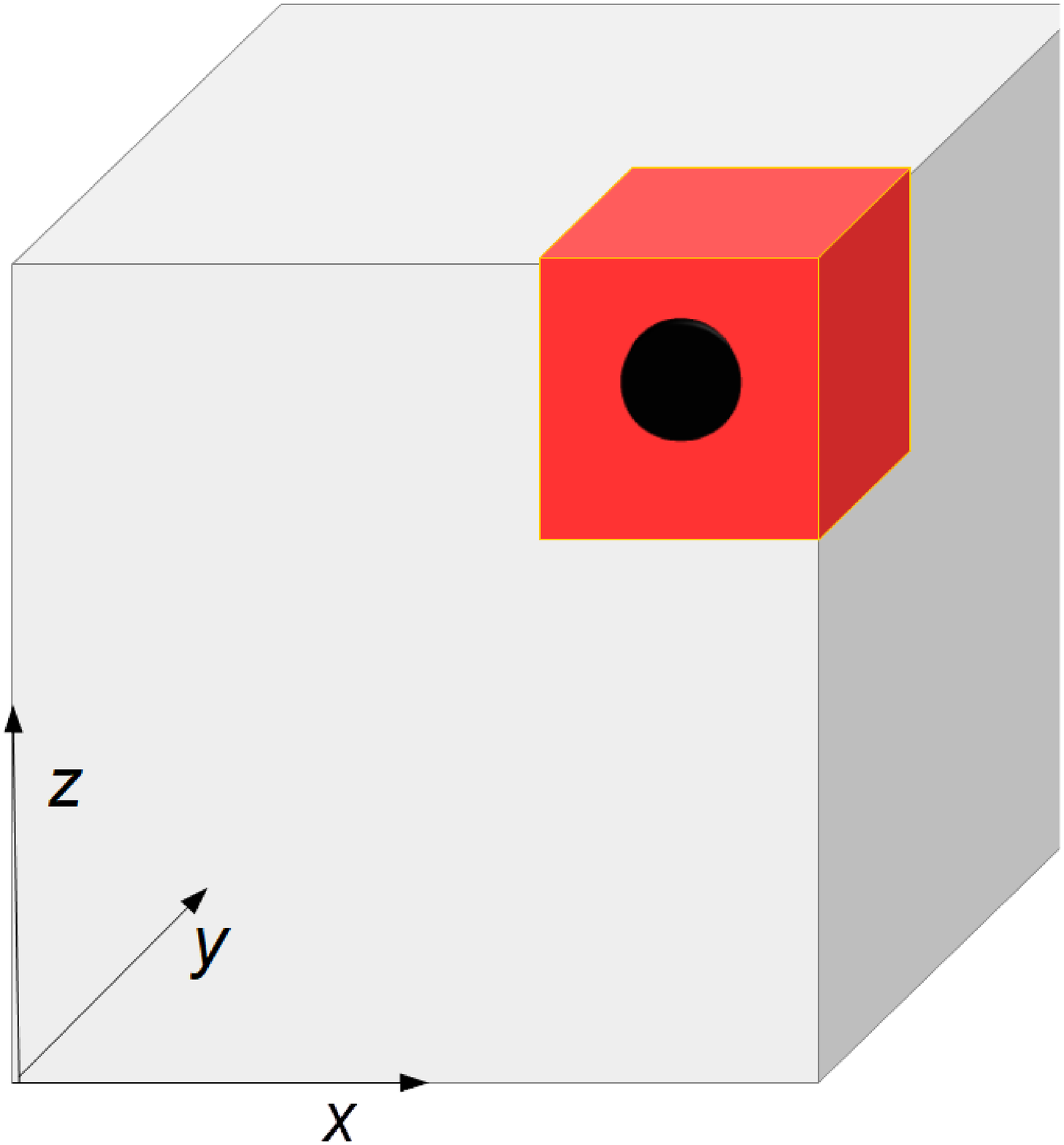}
	\caption{3D criticality threshold}
    \label{vue_seuil_3D}
\end{figure}

The interest and originality are characterized by the possibility of classifying descriptors and thus extending the notion of interval to that of space. Our positioning is also conceivable in a 3D space as shown in Figure \ref{vue_seuil_3D}. The example reveals the positioning of three coordinates specific to three state variables at the instant $t$. Note that the point belongs to a volume typed as "critical". 

In a non-evolving environment made up of a limited number of :
\begin{itemize}
	\item cells (grid 2D) ;
	\item of volumes (3D affine space) ;
	\item species (personnal, infection) ;
\end{itemize} 
a library of thresholds is a possible but restrictive solution.

 We suggest that to overcome this limitation, we couple our MAS with case reasoning. In this way, we lift the scientific lock posed by the rigidity of a complex system without the acquisition of evolutionary and adapted knowledge.  
Indeed, according to Jean Lieber \cite{lieber2008contributions}, "reasoning from cases is to solve the target problem by using a case base, and consequently by adapting to the target problem. :\\
  CBR : (target,CaseBase) $\longmapsto$ Sol(target) $\in$ Solutions "\\
    
The CBR consists of the following 5 main steps: 
    \begin{itemize}
		\item \textbf{elaborate} which consists in formalizing a problem in order to make it exploitable by a machine.	    
	    \item  \textbf{retrieve} which consists in solving the problem from the case base : \\
		rememoration: (Target,CaseBase) $\longmapsto$ (srce,Sol(srce)) $\in$ CaseBase. 
		\item \textbf{reuse} which consists in applying the solution of the problem belonging to the case base and possibly adapting it. The adaptation consists in solving the target solution from the recalled case.  (srce,Sol(srce)).\\
		 Adaptation : (srce,Sol(srce),target). $\longmapsto$ Sol(target) 	
		\item  \textbf{review} allowing the expert to make corrections to the target solution proposed by the system.
		\item \textbf{retain} which consists of enriching the case base with new solutions. 
    \end{itemize}
 
 \paragraph{Elaborate.}
Applied to our situation, the cases are defined by the application: $U \rightarrow R$ with $U$ characterizing a sequence of quadruplets $(E,A,V,t)$ and $R$ the torque (state of the system, recommendation). To each entity $E$, is assigned the attribute $A$ of value $V$ at instant $t$ (cycle). In the example (nurse,fatigue,1.5,1200) $\rightarrow$ (N,Normal), the value 1.5 is assigned to the "fatigue" attribute of the "nurse" agent at cycle 1200. The application returns an "N" state ($\neg$ alert) and a "Normal" recommendation.  
   
 \paragraph{Retrieve.} 
The reminder step, which consists in finding the most similar cases, is based in a first step on a filtering of the source cases which have the same number of quadruplets as the target case. In a second step, a similarity calculation will allow to calculate with a finer granularity the most similar case among those already extracted. As the quadruplets are not necessarily homologous (different attributes), we have chosen to compare each quadruplet of the target case to all the quadruplets of the source case. The chosen vector distance calculation is defined according to the relation : 


\begin{small}
 \[ sim(\overrightarrow{C},\overrightarrow{S}) = \dfrac{1}{n^{2}}  \sum_{i=1}^{n} \sum_{j=1}^{n}  \sqrt{\sum_{k=1}^{4}  \left( 1-I \right)^{2}}\]
\end{small} 

$I=\left( \frac{x_{ki}}{y_{kj}}\right)$ si $x_{ki} \leq 
y_{kj}$. Otherwise ($y_{kj} \leq 
x_{ki}$) $I=\left( \frac{y_{kj}}{x_{ki}}\right)$.
 $x_{ki}$ and $y_{kj}$ are respectively the values associated with each element of the target quadruplet ($\overrightarrow{C}$) and the source quadruplet ($\overrightarrow{S}$) among $n$ quadruplets.  Comparing the elements by quotient them exonerates us from the problems of scales specific to the different descriptors. The column ``similar'' of the fig \ref{tab:cas_sim} gives an example of the result of this calculation.   

\paragraph{Reuse.} 
This step defines the target case. In the case of automatic processing, the inheritance of the source case solution will modify the attributes of the agents belonging to the affine criticality spaces (2D grid, or 3D volume). This update of the SMA database is carried out in parallel with that of the CBR database which is enriched with a new case.   
When an adaptation is necessary, (sufficient but imperfect similarity), we apply the following task decomposition proposed by Jean Lieber \cite{lieber2008contributions} :

\begin{center}
	path(srce,target) = ($p_{0}\overset{r1}{\rightarrow}p_{1}\overset{r2}{\rightarrow}p_{2}...p_{n-1}\overset{rn}{\rightarrow}p_{n}$ )  
\end{center}


As an example, consider the following case: 
\begin{center}
  target $p_{0}$ = (nurse$\_$fatigue$\_$limit$\wedge$nurse$\_$dexterity$\_$threshold), \\
    srce $p_{n}$ = (surgeon$\_$fatigue$\_$limit$\wedge$surgeon$\_$dexterity$\_$limit).\\
\end{center}

\begin{figure}[h!]
   \centering
   \includegraphics[scale=0.40]{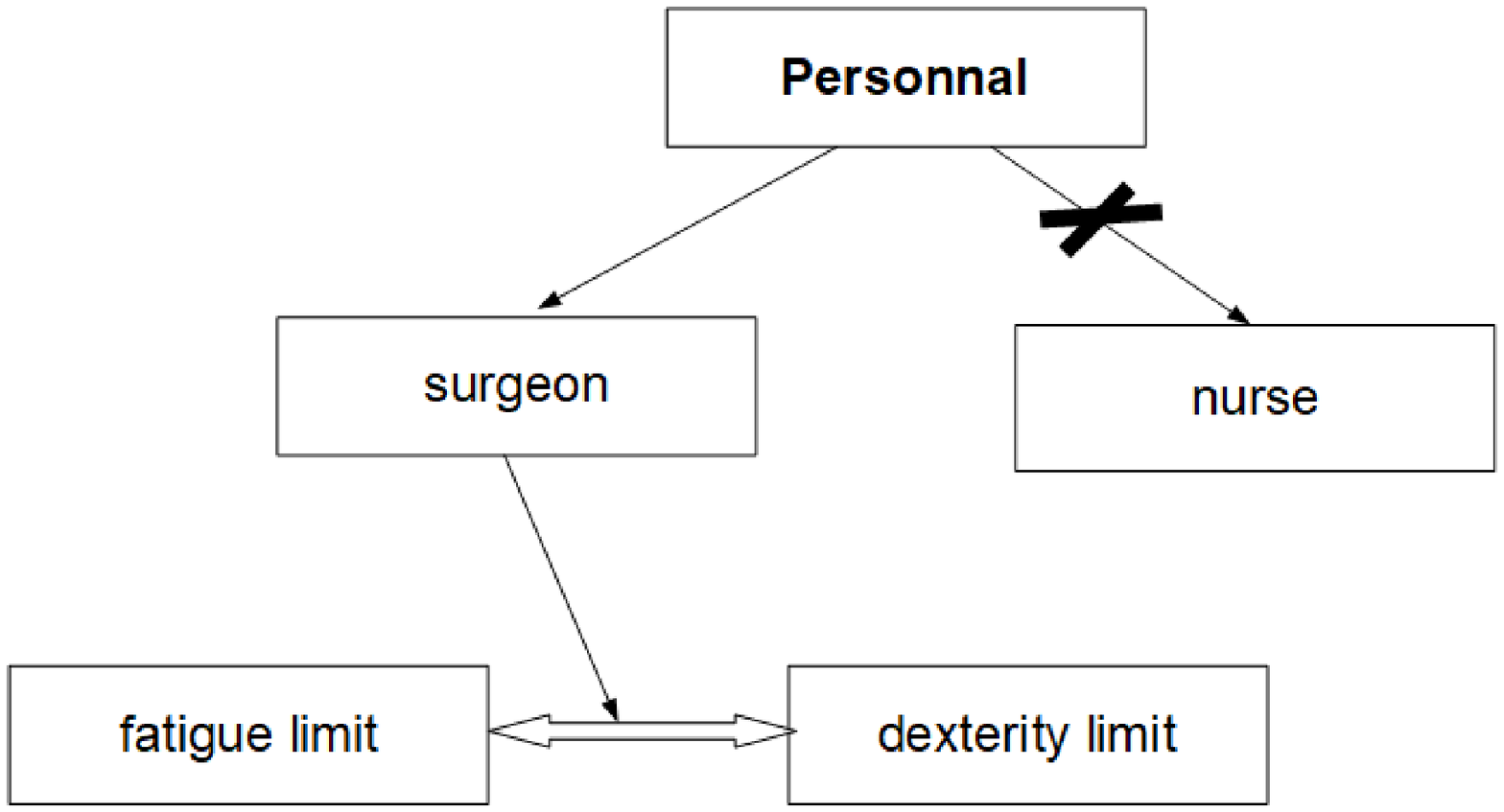}
   \caption{Adaptation}
   \label{adaptation}
\end{figure}

The Figure \ref{adaptation} which summarizes this query shows us that the interviewed agent does not belong to the case base (sucker $\neq$ nurse, but that he belongs to a parent (personal) class with which it is possible to compare the degree of similarity of an agent of the same nature (belonging to the same group). We can thus associate to this new class the collective thresholds inherited from the sucker class. \\

\paragraph{Review and retain.}
Each new target case, after possibly being corrected by experts in the field, is added to the case database and at the same time to the MAS case database with the aim of optimizing the search for similar events. 

\paragraph{Implementation}

The choice of the platform for the development of our simulator was GAMA \cite{amouroux2007gama} which allows to build the models in an Integrated Development Environment (IDE) incorporating the GAMA Modeling Language (GAML). This specificity guarantees the enrichment of the model during its implementation. The platform, which is rich in several components, also allows several visualization models to be placed in a display window. Finally, it is possible to build very complex models thanks to high-performance space management tools in different synchronized environments within a continuous reference space.

The following section presents the first results obtained
in a simulation context based on the risk of fatigue
infectious and human.

\section{Results}
\label{sec:result}

The first results obtained concern alerts of human fatigue and infectious spread. Alert levels are defined for each agent according to their objectives, roles and attributes. The following paragraphs describe the results obtained for each of these agents.

\paragraph{Fatigue modeling}
~\\

\begin{figure}[h!]
	\centering
	\includegraphics[scale=0.55]{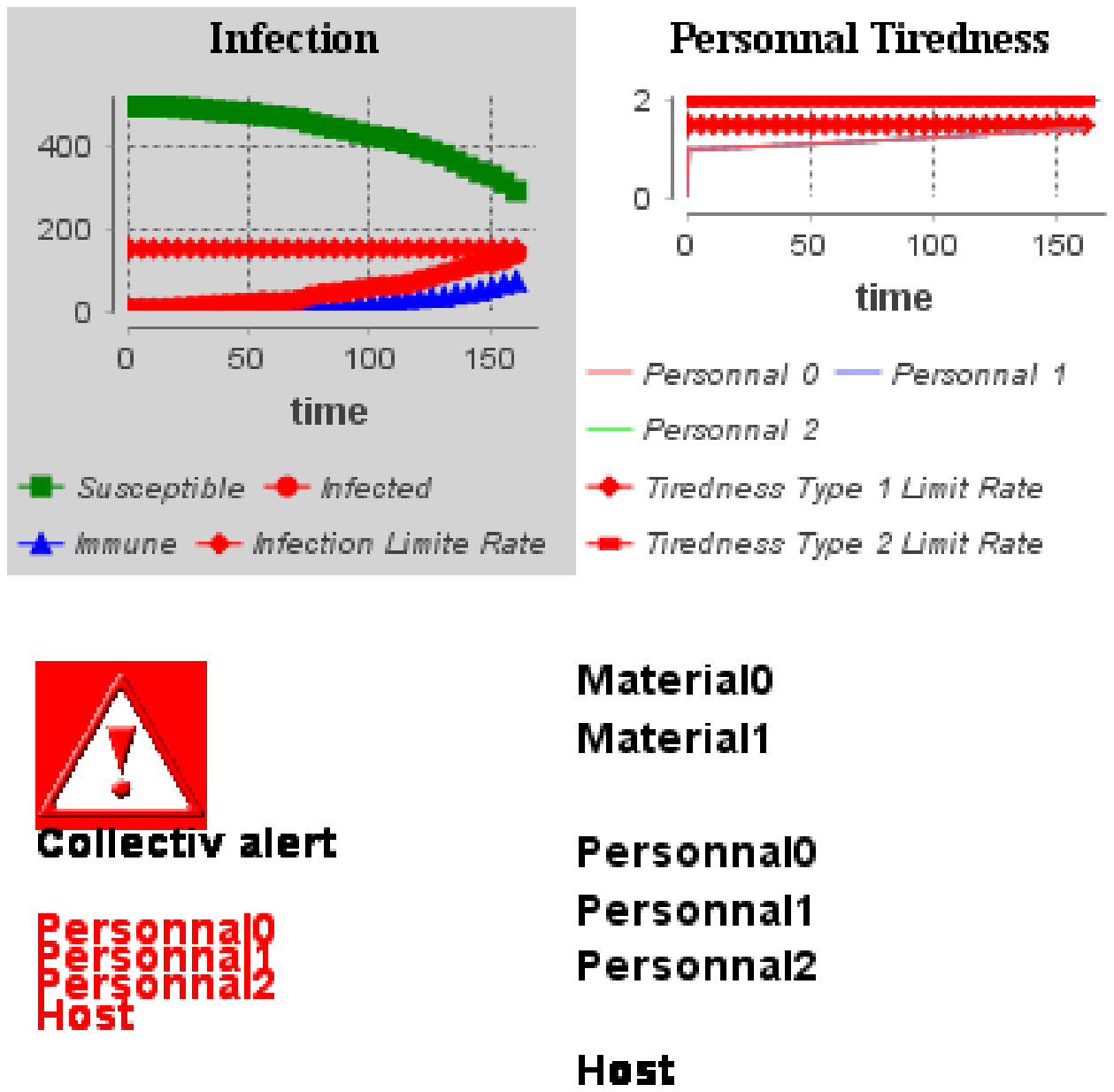}
	\caption{Coupling infection and fatigue}
	\label{couplage_infection_fatigue}
\end{figure}

The human fatigue curve observed in the "Personnal Tiredness" graph in Figure \ref{couplage_infection_fatigue} is a function equivalent to a hyperbole whose evolution depends on the type defined in the parameter (sleep, anxiety ...). In our example we have considered only two individuals knowing that it is possible to assign a type of fatigue to each person working in the operating room. The alert threshold is defined by the user.

\paragraph{Infection modeling.}
~\\
Observable results for infectious risk show a correlation between the evolution of contaminating agents and the decrease in the healthy population. In addition, the interaction of immune agents can reverse this trend. The autonomous and random displacement in space of these entities makes it possible to consider all the scenarios according to the parameters initially defined and modifiable during the simulation. The input data are statistical data and it is quite possible to obtain them with real-time capture by integrating the Big Data. Like fatigue, the alert threshold is user-defined. 

\paragraph{Simulation of the risk of fatigue coupled with the risk of infection: an MAS approach}
~\\
One of the advantages of our multi-agent approach is the ability to couple species as shown in the "Collectiv alert" in Figure \ref{couplage_infection_fatigue} where the risks "fatigue" and "infection" are aggregated. A threshold is defined by the user knowing that the value of the risk is a combined calculation with all other risks. In our example, it is reached while the infection threshold is not. However, a restriction appears when the attributes to be compared require a large number of combinations. Thus, the aggregation of four agents would require $2^{4}$ combinations (all parts of a set). On the other hand, the learning of the MAS, is dependent on capitalized experiences. 
\paragraph{Simulation of Fatigue Risk Associated with Risk of Infection: A MAS/CBR Approach}
~\\
In this approach, the limit thresholds do not result from a priori parameterization, but are derived from a case base enriched by experience. The simulations feed this base "as the water goes" according to a predefined amplitude (every 100 cycles in our example). During this learning phase, the search for similarities (recall stage of the CBR  cycle) generates a number of eligible cases which are then compared to the target case and possibly adapted. Continuous adjustment by the experts guarantees the efficiency of the results. 

\begin{figure*}[h!]
  \centering
\begin{tabular}{|c|c|c|c|c|c|c|c|c|c|c|}
  \hline 
  \rule[-1ex]{0pt}{2.5ex}  Case & E & A & V & t & E & A & V & t & Simil & Alert/ \\ 
  \rule[-1ex]{0pt}{2.5ex}  Num. &  &  &  &  &  &  &  & &  & Preco. \\
  \hline 
  \rule[-1ex]{0pt}{2.5ex} 0 & surgeon & fatigue & 2.7 & 400 & staphy & infection & 270 & 400 &  &  \\ 
  \rule[-1ex]{0pt}{2.5ex} (cible) &  &  &  &  &  &  &  &  &  &  \\
  \hline 
  \rule[-1ex]{0pt}{2.5ex} 1 & surgeon & fatique & 3.2 & 700 &  &  &  &  &  &  \\ 
  \hline 
  \rule[-1ex]{0pt}{2.5ex} 8 & bistoury & fatigue & 0.9 & 1200 & nurse & fatigue & 2.1 & 1200 & 1.482 & N/Normal \\  
  \hline 
  \rule[-1ex]{0pt}{2.5ex} 35 & nurse & fatique & 2.5 & 300 & staphy & infection & 280 & 300 & 1.018 & O/Pause \\  
  \rule[-1ex]{0pt}{2.5ex} &  &  & & & & & & & & Pers. \\
  \hline 
 \end{tabular}   
  \caption{Recherche de cas similaires}
  \label{tab:cas_sim}
\end{figure*}

Figure \ref{tab:cas_sim} partially summarizes (not all cases are represented) a case search. Among the cases in the database with 2 quadruplets, case 35 is the best candidate. The proximity of the case ($1.018 \ll 1.2$ (configurable acceptance threshold) generates the adaptation: IBODE $\rightarrow$ CHIR and returns to the user the state of the system (alert) and a recommendation (surgical agent break). The threshold of 1.2 was chosen because a similarity is verified between the target case and the source case in 95 \% of the cases when $sim(\overrightarrow{C},\overrightarrow{S})\leq 1.2$. The memorization which follows allows on the one hand the enrichment of the case base and on the other hand the update of the knowledge base of the MAS agents. 

\begin{figure}[h!]
	\centering
	\includegraphics[scale=.70]{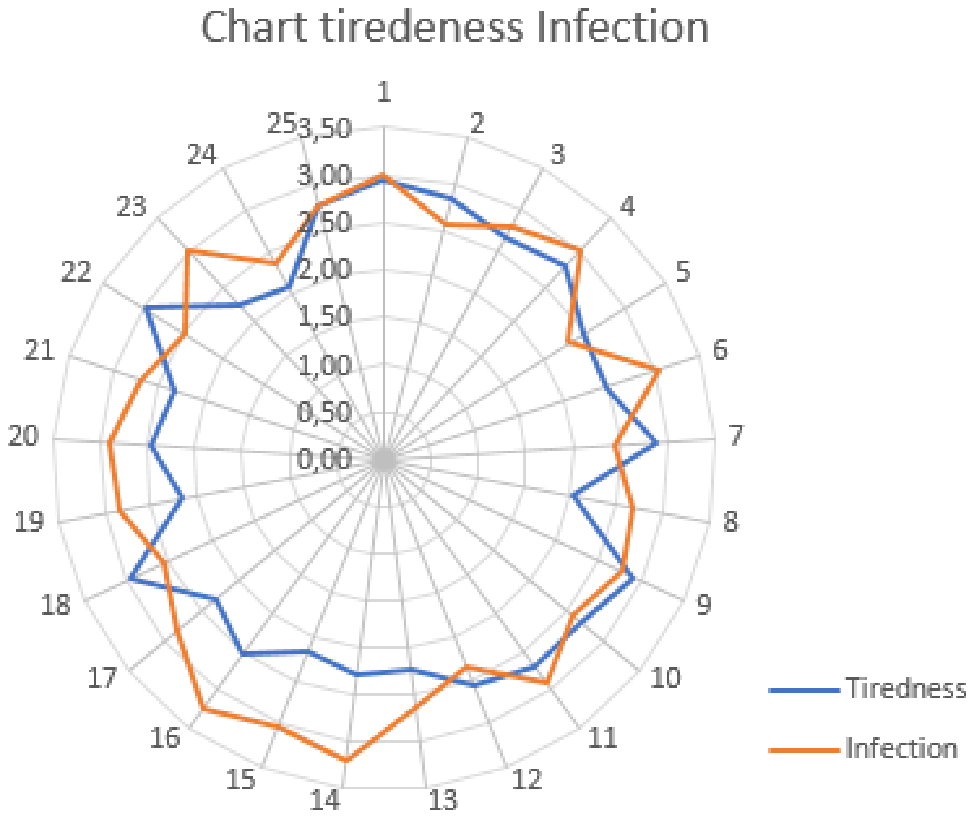}
	\caption{Graph 25 simulations coupling infection and fatigue}
    \label{diagramme_simulation_fatigue_infection}
\end{figure}

\paragraph{Agent Alert}
~\\
The "alert" centralizing agent integrates all risks, whether individual or collective. Safety in the operating room is no longer just a set of individual but also collective alerts. This micro and macro approach makes it possible to envisage critical levels that are invisible to date. Indeed, the alert observed in our example is reached whereas it was not individually reached for the agents.
\\
In the diagram in Figure \ref{diagramme_simulation_fatigue_infection} we have represented the result of 25 simulations of the risk of infection coupled with human fatigue. The values express the triggering of the alert resulting from the aggregation of the level of human fatigue coupled with that of the infectious risk as a function of time (number of cycles). These collective alerts are triggered when individual thresholds are not reached.  For the sake of clarity, 25 simulation values were chosen instead of 100, however, the trends of the curve are similar and show us that the results vary little when the parameters are identical.  

\section{Discussion}
\label{sec:discut}

The multi-agent paradigm allows for multi-level modeling and allows us to consider the same character at several scales. Thus, for example, we can distinguish within the operating theatre the infection of an organ (macro) but also the quantity of infectious particles present in the air (micro). Concerning predictivity, the agentification of criticality zones has made it possible to refine the level of thresholds. The first simulations reveal a likelihood of results with similar curve trends. The standard deviation of less than 0.5 on all the values confirms this assessment, thus making it possible to envisage the implementation of responses in terms of safety. However, the simulated data will have to be compared with the actual data in a future study.

However, limits are observable when the number of quadruplets to be compared is greater than three. Indeed we lose the fineness of the location offered by 2D or 3D affine spaces. The MAS/CBR coupling is a promising answer. Currently, we consider the case base as an endogenous resource to CBR which would deserve to be extended to other knowledge acquisition systems such as trace bases or some external bases. 

In the next step, together with the medical team, we will define the data for the model and comment on the results. This is an important phase both in terms of validating the results and in terms of development. It is during this phase that we will determine, for example, other risks to be modeled, knowing that the only limits are material ones. We will also define the data acquisition mode. This will allow the system to use the data online dynamically or offline. In the first case, we will consider the simulator as an alert generator, and in the second case as a scenario generator for predictive purposes.

\section{Conclusion}

In this article, we presented the simulator we designed and implemented to predict the risks associated with human and material fatigue and hospital-acquired infections in surgery. The results proved that our model can handle such data in a multi-scale environment. Similarly, our multidimensional approach to criticality levels has opened up many perspectives in terms of optimizing results (granularity). The limits observed beyond a 3D space have opened our field of investigation towards case-based reasoning that optimizes the knowledge acquisition of our simulator. Our system also proved its efficiency since most of the simulated situations were correct. Nevertheless, in a second phase, we plan to integrate our system in dynamic mode (connected to the sensors) and to design a module of corrective proposals. 

\section*{Acknowledgements}

The authors would like to thank Prof. Auber and Dr. Boulard from the Department of Pediatric Surgery of the Centre Hospitalier R{\'e}gional Universitaire de Besan\c{c}on for their expertise in the field of application of the study. 

\bibliography{jfsma_va}

\end{document}